\documentclass{article}

\usepackage{microtype}
\usepackage{graphicx}
\usepackage{subcaption}
\usepackage{tikz}
\usetikzlibrary{positioning,shapes.geometric,shapes.symbols}
\usepackage{booktabs} % for professional tables
\usepackage{multirow} % for multi-row table cells
\usepackage{hyperref}
\makeatletter
\g@addto@macro\UrlBreaks{\do\/\do-\do\_\do\.\do\?\do\&\do\=\do\#\do\%}
\makeatother
\usepackage[preprint]{icml2026}
\usepackage{amsmath}
\usepackage{amssymb}
\usepackage{mathtools}
\usepackage{amsthm}
\usepackage{placeins} % in preamble

\usepackage[capitalize,noabbrev]{cleveref}

\theoremstyle{plain}

\theoremstyle{definition}

\theoremstyle{remark}

\usepackage[textsize=tiny]{todonotes}

\icmltitlerunning{Don't Break the Cache: Prompt Caching for Long-Horizon Agentic Tasks}

\begin{document}

\twocolumn[
  \icmltitle{Don't Break the Cache: An Evaluation of Prompt Caching\\for Long-Horizon Agentic Tasks}



  \begin{icmlauthorlist}
    \icmlauthor{Elias Lumer}{pwc}
    \icmlauthor{Faheem Nizar}{pwc}
    \icmlauthor{Akshaya Jangiti}{pwc}
    \icmlauthor{Kevin Frank}{pwc}
    \icmlauthor{Anmol Gulati}{pwc}\\
    \icmlauthor{Mandar Phadate}{pwc}
    \icmlauthor{Vamse Kumar Subbiah}{pwc}
  \end{icmlauthorlist}

  \icmlaffiliation{pwc}{PricewaterhouseCoopers, U.S.}

   \vskip 0.3em
  {\centering\normalsize\itshape PricewaterhouseCoopers, U.S.\par}
  
  %{\centering\normalsize\itshape PricewaterhouseCoopers, U.S.\par}

  \icmlcorrespondingauthor{Elias Lumer}{elias.lumer@pwc.com}

  \icmlkeywords{Prompt Caching, Large Language Models, Agentic AI, Tool Calling, API Optimization}

  \vskip 0.3in
]

\printAffiliationsAndNotice{}

\begin{abstract}
    Recent advancements in Large Language Model (LLM) agents have enabled complex multi-turn agentic tasks requiring extensive tool calling, where conversations can span dozens of API calls with increasingly large context windows. However, although major LLM providers offer prompt caching to reduce cost and latency, its benefits for agentic workloads remain underexplored in the research literature. To our knowledge, no prior work quantifies these cost savings or compares caching strategies for multi-turn agentic tasks. We present a comprehensive evaluation of prompt caching across three major LLM providers (OpenAI, Anthropic, and Google) and compare three caching strategies, including full context caching, system prompt only caching, and caching that excludes dynamic tool results. We evaluate on DeepResearch Bench, a multi-turn agentic benchmark where agents autonomously execute real-world web search tool calls to answer complex research questions, measuring both API cost and time to first token (TTFT) across over 500 agent sessions with 10,000-token system prompts. Our results demonstrate that prompt caching reduces API costs by 41-80\% and improves time to first token by 13-31\% across providers. We find that strategic prompt cache block control, such as placing dynamic content at the end of the system prompt, avoiding dynamic traditional function calling, and excluding dynamic tool results, provides more consistent benefits than naive full-context caching, which can paradoxically increase latency. An ablation study across prompt sizes (500-50,000 tokens) and tool call counts (3-50) demonstrates universal linear cost and TTFT benefits, after the provider caching token minimum, and reveal provider-specific strategy discrepancies across variants. We provide nuanced discussion and guidance for implementing prompt caching in production agentic systems.
\end{abstract}

\section{Introduction}
\label{sec:introduction}

Recent advancements in Large Language Model (LLM) agents have enabled complex, long-horizon agentic tasks that require extensive tool calling across multi-turn conversations \citep{ji2025manus_context_engineering}. Through function calling, LLM agents can invoke APIs, execute web searches, interact with databases, and perform domain-specific actions on behalf of users. As these agentic workloads grow in complexity, conversations can span dozens of API calls with context windows accumulating tens of thousands of tokens, leading to significant costs and latency overhead. To address this, major LLM providers including OpenAI, Anthropic, and Google offer prompt caching, a feature that reuses previously computed key-value (KV) tensors from attention layers to avoid redundant computation on repeated prompt prefixes \citep{openai_prompt_caching_docs, anthropic_prompt_caching_docs, google_vertex_context_cache_overview}.

\begin{figure*}[t]
\centering
\includegraphics[width=10.5cm]{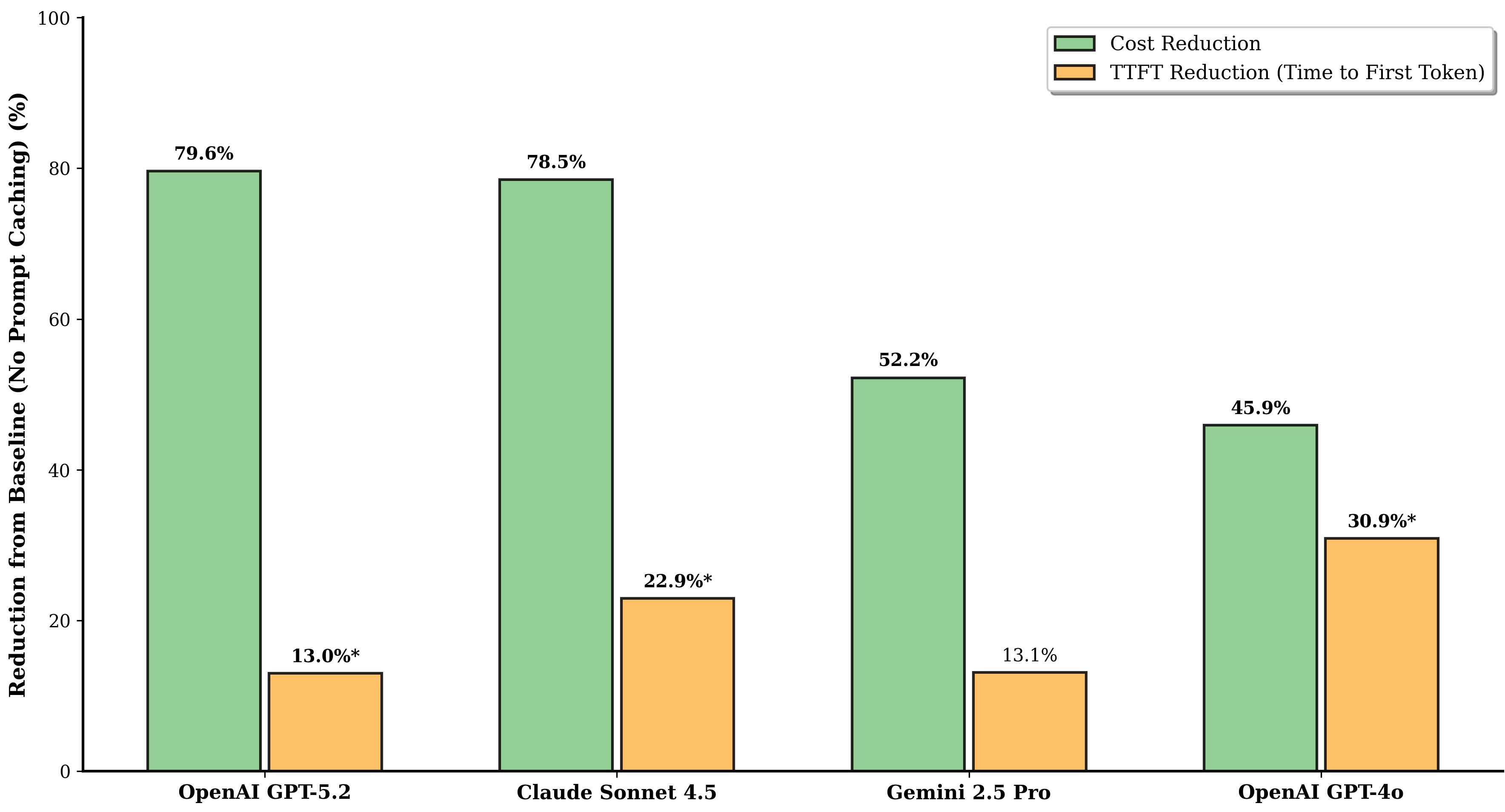}
\caption{
Prompt caching benefits (best cache mode per model).
Percentage reduction in API cost and time to first token (TTFT) relative to a no-cache baseline.
Asterisks denote statistically significant TTFT improvements ($p < 0.05$).
}

\label{fig:summary_results}
\end{figure*}

While providers offer reduced pricing for cached input tokens, the benefits of prompt caching in real-world agentic workloads remain under-explored in the research literature. Existing work on KV cache optimization focuses primarily on inference-level memory management and compression \citep{kwon2023pagedattention, ge2023adaptivekvcache, shi2024keepthecostdown}, rather than evaluating the enterprise-grade prompt caching features offered through provider APIs. Concurrent work has audited prompt caching across providers to detect timing side-channel vulnerabilities \citep{gu2025auditingpromptcachinglanguage}, but to our knowledge, no prior work has quantified the cost benefits of prompt caching or compared caching strategies for agentic workloads. This gap is particularly significant given the recent proliferation of long-running agents for deep research, coding assistance, and autonomous task completion, where prompt caching could substantially reduce operational costs and improve user experience through faster response times.

In this paper, we present the first comprehensive evaluation of prompt caching strategies for long-horizon agentic tasks across three LLM providers (OpenAI, Anthropic, and Google) using four flagship models (Figure~\ref{fig:summary_results}). We compare three caching strategies, including full context caching, system prompt only caching, and caching that excludes dynamic tool results. We evaluate on DeepResearch Bench \citep{du2025deepresearchbenchcomprehensivebenchmark}, a multi-turn agentic benchmark where agents autonomously execute web search tool calls to answer research questions. Our evaluation spans 500 agent sessions with 10,000-token system prompts, measuring both API cost and time to first token (TTFT) across all conditions.

Our evaluation reveals three key findings:

\textbf{Prompt caching delivers substantial and consistent cost savings across all providers:} All four models tested show statistically significant cost reductions when prompt caching is enabled. Cost savings range from 41\% to 80\% across providers. These savings are consistent across all three caching strategies, demonstrating that prompt caching provides reliable cost benefits regardless of the specific caching approach employed.

\textbf{Latency improvements vary significantly across providers and require careful strategy selection:} Time to first token improvements range from 13\% to 31\% across providers, though latency variance differs substantially between providers. Notably, the cache strategy that maximizes cost savings does not always maximize latency improvement, highlighting the importance of strategy selection based on optimization goals.

\textbf{Strategic cache boundary control outperforms naive full-context caching:} Providers abstract much of the caching mechanism, automatically triggering cache creation when token thresholds are exceeded. However, naively enabling full-context caching can paradoxically increase latency, as dynamic tool calls and results may trigger cache writes for content that will not be reused across sessions. By strategically controlling cache boundaries, such as caching only the system prompt or explicitly excluding tool results, practitioners can ensure that only stable, reusable content is cached. Our results show that system prompt only caching provides the most consistent benefits across both cost and latency dimensions.

\section{Background}
\label{sec:background}

\subsection{KV Cache and LLM Inference}
\label{subsec:kv_cache}

Large Language Model inference consists of two distinct phases: the prefill phase, where the model processes the input prompt and generates attention key-value (KV) tensors, and the decode phase, where the model autoregressively generates output tokens \citep{pope2022efficiently}. During prefill, the model computes attention over the entire input sequence, producing KV tensors that capture the contextual representations needed for subsequent generation. These KV tensors are stored in the KV cache and reused during decoding to avoid redundant computation, enabling efficient token-by-token generation \citep{notlain2025kvcaching}.

%As context windows have grown from thousands to millions of tokens, KV cache management has become a critical bottleneck in LLM serving \citep{shi2024keepthecostdown}. The memory footprint of KV caches scales linearly with sequence length and batch size, often consuming more GPU memory than the model weights themselves for long-context workloads. This has motivated extensive research on KV cache optimization, including memory management techniques such as PagedAttention \citep{kwon2023pagedattention}, which applies paging-style memory management to reduce fragmentation and waste, achieving 2-4x throughput improvements. Other approaches focus on KV cache compression through selective retention of important tokens \citep{ge2023adaptivekvcache}, storage-compute tradeoffs that balance recomputation against cache loading \citep{jin2024cake}, and shared prefix optimization for high-throughput inference \citep{juravsky2024hydragen, wu2024layer, shadowkv2024, survey_efficient_inference2024}.

This has motivated extensive research on KV cache optimization, including memory management techniques such as PagedAttention \citep{kwon2023pagedattention}, which applies paging-style memory management to reduce fragmentation and waste. Other approaches focus on KV cache compression through selective retention of important tokens \citep{ge2023adaptivekvcache}, storage-compute tradeoffs \citep{jin2024cake}, and shared prefix optimization for high-throughput inference \citep{juravsky2024hydragen, wu2024layer, shadowkv2024, survey_efficient_inference2024}.

\subsection{Prompt Caching in Provider APIs}
\label{subsec:prompt_caching}

While KV caching is a general inference optimization technique, prompt caching refers to the productized, provider-managed features that reuse KV tensors across API requests when prompts share common prefixes \citep{openai_prompt_caching_docs, Gim2024}. By caching the KV tensors from the prefill phase, providers can skip redundant computation when subsequent requests begin with the same content, reducing both latency and cost for users.

Major LLM providers have implemented prompt caching with varying approaches. OpenAI offers automatic prompt caching on GPT-4o and newer models, where caching activates automatically for prompts exceeding a minimum token threshold, with cache hits occurring only for exact prefix matches \citep{openai_prompt_caching_docs, openai_prompt_caching_blog_2024}. Anthropic provides developer-controlled caching through explicit cache breakpoints, allowing users to specify which portions of their prompt should be cached, with configurable time-to-live (TTL) options \citep{anthropic_prompt_caching_docs, anthropic_prompt_caching_blog_2025}. Google offers both implicit caching, which activates automatically with no guaranteed cost savings, and explicit context caching, where developers create and reference caches with guaranteed discounts \citep{google_vertex_context_cache_overview, google_gemini_implicit_caching_blog_2025}.

Implementation details such as minimum token thresholds (typically 1,024-4,096 tokens depending on model, see Table~\ref{tab:caching_minimums}), TTL durations (ranging from 5 minutes to 24 hours), and pricing structures vary across providers and are subject to change \citep{prompthub_prompt_caching_2025, Azure2025}. These differences have practical implications for cache hit rates and cost optimization. Recent work has audited prompt caching across 17 providers, demonstrating that cache hits produce measurable TTFT reductions and identifying security vulnerabilities from timing side-channels \citep{gu2025auditingpromptcachinglanguage}. However, their focus on security auditing using synthetic prompts and smaller, older generation models does not compare caching strategies or evaluate cost and latency benefits for long-running agentic tasks on modern flagship models.

\subsection{Agentic Workloads and Context Engineering}
\label{subsec:agentic_workloads}

Recent advances in LLM agents have enabled complex, long-horizon tasks that extend far beyond single-turn question answering. Modern agentic applications including deep research assistants, coding agents such as Claude Code and Cursor, and autonomous task completion systems like Manus routinely execute 30-50 or more tool calls within a single session \citep{ji2025manus_context_engineering, du2025deepresearchbenchcomprehensivebenchmark, mialon2023gaia, zhou2023webarena, drouin2024workarena, wei2025browsecomp}. In such workflows, each tool call adds content to the conversation context, including the tool invocation, execution results, and the model's subsequent reasoning, causing context windows to grow rapidly throughout the session.

This growth presents challenges for prompt caching \citep{multiturn_survey2025,laban2025llmslostmultiturnconversation}. Recent work has proposed solutions for multi-turn caching scenarios \citep{flashgen2025, yan2025contextcachecontextawaresemanticcache}. Unlike static question-answering scenarios where prompts are largely predetermined, agentic workloads feature dynamic, session-specific content that accumulates unpredictably. Tool results often contain user-specific data that will not benefit other sessions, and the interleaving of static system prompts with dynamic tool outputs complicates cache reuse. Context engineering strategies have emerged to manage these challenges, including treating external storage as extended memory and structuring prompts to maximize cache efficiency \citep{ji2025manus_context_engineering,lumer2025toolagentselectionsurvey}. However, the effectiveness of prompt caching across different caching strategies in agentic workloads has not been comprehensively evaluated. Our work addresses this gap by measuring cost and latency benefits across controlled caching strategies on a multi-turn agentic benchmark.

\section{Methodology}
\label{sec:methodology}

\subsection{Experimental Setup}
\label{subsec:experimental_setup}

We evaluate prompt caching across three major LLM providers: OpenAI, Anthropic, and Google. For each provider, we select flagship models: GPT-5.2 from OpenAI, Claude Sonnet 4.5 from Anthropic, and Gemini 2.5 Pro from Google. We additionally include GPT-4o to examine caching behavior across model generations. All selected models support prompt caching through their respective APIs.

We use DeepResearch Bench \citep{du2025deepresearchbenchcomprehensivebenchmark} as our evaluation benchmark, a multi-turn agentic benchmark where agents autonomously execute web search tool calls to answer complex research questions. We selected this benchmark over alternatives such as other deep research benchmarks \citep{bosse2025deepresearchbench_futuresearch, li2025reportbench} due to its focus on tool-intensive agentic workflows and real-world 100 PhD-level research tasks, each crafted by domain experts across 22 fields. We implement our research agent using Deep Agents \citep{langchain2025deepagents}, one of various open source libraries for creating long-running agents \cite{anthropic2025claudeagentsdk,openai2025agentssdk,google2025adk,microsoft2023autogen,microsoft2026agentframework,microsoft2023semantickernel,crewai2023crewai,llamaindex2022llamaindex,huggingface2024smolagents,openai2024swarm,agno2026agno}. Each agent session begins with a research question and the agent iteratively calls a web search tool to gather information before synthesizing a comprehensive response. This benchmark reflects realistic agentic workloads where context windows grow dynamically through tool invocations and results.

For each model, we conduct 40 independent agent sessions per cache condition, with each session answering a unique research question from the benchmark. Sessions use a 10,000-token system prompt containing agent instructions for deep research, including guidance on tool usage, question decomposition, and report synthesis. Each session starts with a fresh context, ensuring that cache benefits are measured within individual multi-turn conversations rather than across sessions.

\begin{figure*}[t]
\centering
\includegraphics[width=13.5cm]{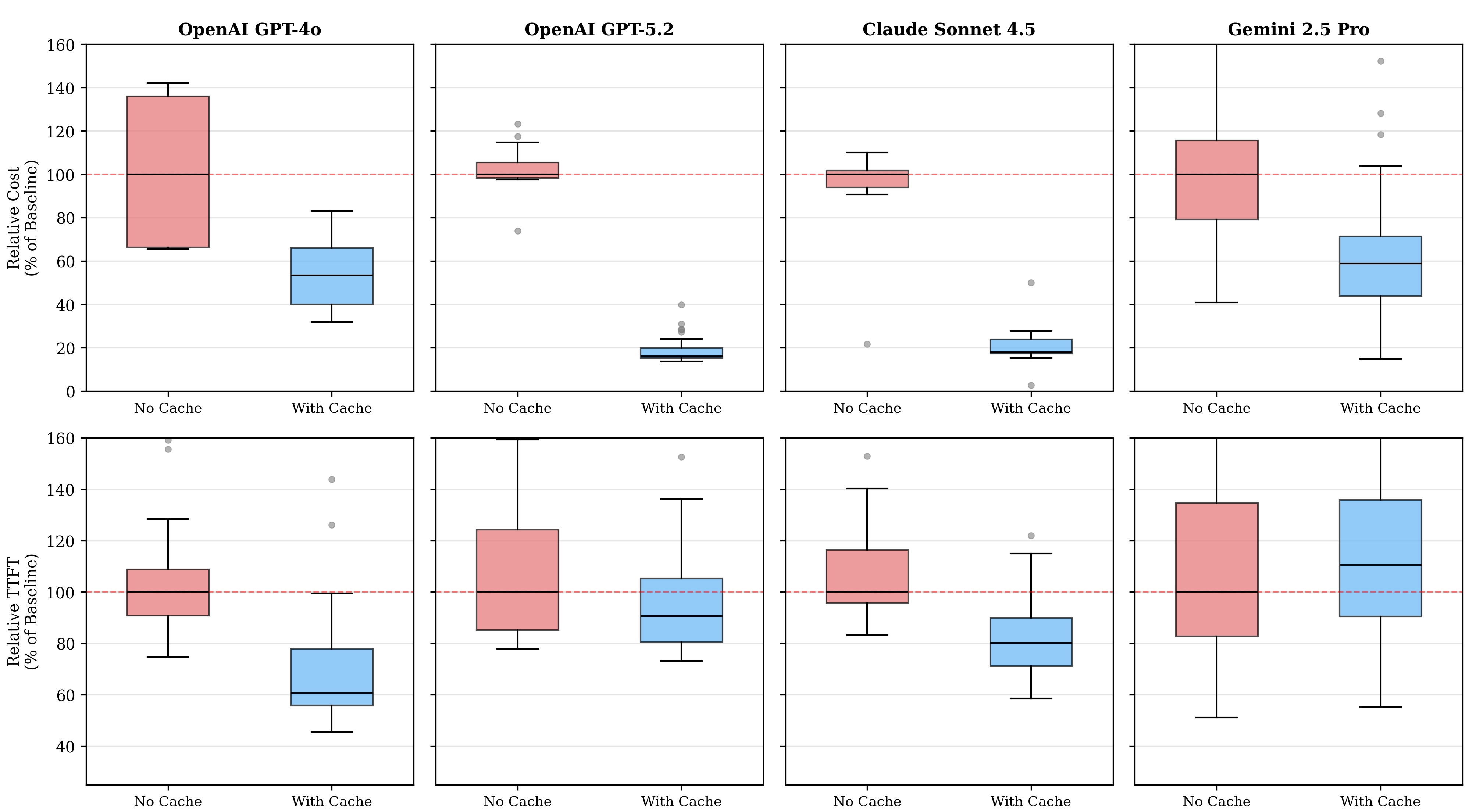}
\caption{
Prompt caching impact for normalized cost and time to first token (TTFT).
Results use the system prompt only caching strategy.
The no-cache baseline is normalized to 100\% and lower values indicate better performance.
}

\label{fig:cost_ttft_comparison}
\end{figure*}

\subsection{Cache Mode Implementation}
\label{subsec:cache_modes}

We implement four cache conditions to systematically evaluate prompt caching strategies (see Appendix~\ref{app:cache_strategies}, Figures~\ref{fig:no_cache_strategy}--\ref{fig:exclude_tool_results_strategy} for visual illustrations). To control cache boundaries precisely, we use unique identifiers (UUIDs) to break the cache at specific points in the prompt, ensuring that content after the UUID is not cached from previous requests.

\textbf{No Cache (Baseline):} A UUID is prepended to the beginning of the system prompt, breaking the cache immediately and forcing the model to recompute all tokens. This serves as our baseline condition where no caching benefits are realized. In real-world agentic tasks, this symbolizes including dynamic content, such as timestamps and user information, to the system prompt on inference time \cite{ji2025manus_context_engineering}.

\textbf{Full Context Caching:} No UUIDs are added, allowing the provider's caching mechanism to operate automatically. OpenAI and Google enable prompt caching automatically for eligible requests, while Anthropic requires explicit cache breakpoints in the API request. This condition represents naive caching where practitioners enable the feature without additional optimization.

\textbf{System Prompt Only Caching:} A UUID is appended to the end of the system prompt, breaking the cache at this boundary. This ensures that only the static system prompt is cached, while the dynamic conversation history, tool calls, and tool results are recomputed on each request.

\textbf{Exclude Tool Results Caching:} UUIDs are appended both after the system prompt and after each tool result. This strategy ensures that tool results, which are dynamic and session-specific, do not contribute to the cache. We found this dual-UUID approach necessary because provider-level KV cache handling can vary, and explicit boundaries provide more predictable caching behavior.

\subsection{Evaluation Protocol}
\label{subsec:evaluation_protocol}

We measure two primary metrics across all conditions: API cost and time to first token (TTFT).

\textbf{Cost:} We calculate cost using token counts reported in API responses, distinguishing between standard input tokens, cached input tokens (cache reads), and cache creation tokens (cache writes). Each token type is multiplied by the corresponding provider pricing (see Appendix~\ref{app:pricing}, Table~\ref{tab:caching_pricing}) to compute total cost per session. Cost is aggregated across all API calls within a session.

\textbf{Time to first token (TTFT):} We measure TTFT using streaming responses, recording the time from request initiation to receipt of the first response chunk. TTFT captures the latency improvement from skipping prefill computation on cached tokens, making it the most relevant latency metric for prompt caching evaluation.

Prior to each experimental condition, we execute warmup calls to prime the cache and record cache creation tokens separately from evaluation runs. Between conditions for different cache modes, we wait sufficient time (exceeding 24 hours) to ensure cache entries expire based on provider TTL policies, preventing cross-condition cache contamination.

\subsection{Statistical Analysis}
\label{subsec:statistical_analysis}

We compare each cache condition against the no-cache baseline using independent samples t-tests. Statistical significance is determined at $\alpha = 0.05$. For each model and cache mode, we report mean cost, mean TTFT, percentage improvement over baseline, and p-values. Sample sizes are $n = 40$ per condition for all models.

\section{Results}
\label{sec:results}

\subsection{Overall Results}
\label{subsec:overall_results}

Table~\ref{tab:summary_results} summarizes the prompt caching benefits across all four models using the best-performing cache mode for each model. All experiments show statistically significant improvements ($p < 0.05$). Cost savings range from 41\% to 80\% across models, while time to first token improvements range from 6\% to 31\%.

\begin{table}[t]
\caption{Prompt caching benefits using the best-performing cache mode for each. Cost savings and TTFT improvement are relative to the no-cache baseline. Bold indicates highest value per metric.}
\label{tab:summary_results}
\centering
\small
\begin{tabular}{llcc}
\toprule
\textbf{Model} & \textbf{Cache Mode} & \textbf{Cost $\downarrow$} & \textbf{TTFT $\downarrow$} \\
\midrule
OpenAI GPT-5.2 & Excl. Tool Results & \textbf{79.6\%} & 13.0\% \\
Claude Sonnet 4.5 & System Prompt & 78.5\% & 22.9\% \\
Gemini 2.5 Pro & System Prompt & 41.4\% & 6.1\% \\
OpenAI GPT-4o & System Prompt & 45.9\% & \textbf{30.9\%} \\
\bottomrule
\end{tabular}
\end{table}

\subsection{Cost Reduction}
\label{subsec:cost_results}

\begin{table*}[t]
\caption{Full comparison of cache modes across all models. Cost savings and TTFT improvement are relative to the no-cache baseline. Negative TTFT values indicate regression.}
\label{tab:full_comparison}
\centering
\small
\begin{tabular}{llcc}
\toprule
\textbf{Model} & \textbf{Cache Mode} & \textbf{Cost $\downarrow$} & \textbf{TTFT $\downarrow$} \\
\midrule
\multirow{4}{*}{OpenAI GPT-5.2} & No Cache (Baseline) & --- & --- \\
& Full Context & 79.3\% & 9.5\% \\
& System Prompt & \textbf{81.4\%} & 10.5\% \\
& Excl. Tool Results & 79.6\% & 13.0\% \\
\midrule
\multirow{4}{*}{Claude Sonnet 4.5} & No Cache (Baseline) & --- & --- \\
& Full Context & 77.8\% & 21.8\% \\
& System Prompt & 78.5\% & 22.9\% \\
& Excl. Tool Results & 78.1\% & 20.9\% \\
\midrule
\multirow{4}{*}{Gemini 2.5 Pro} & No Cache (Baseline) & --- & --- \\
& Full Context & 38.3\% & 6.0\% \\
& System Prompt & 41.4\% & 6.1\% \\
& Excl. Tool Results & 27.8\% & -2.9\% \\
\midrule
\multirow{4}{*}{OpenAI GPT-4o} & No Cache (Baseline) & --- & --- \\
& Full Context & 47.8\% & -8.8\% \\
& System Prompt & 45.9\% & \textbf{30.9\%} \\
& Excl. Tool Results & 46.8\% & 28.1\% \\
\bottomrule
\end{tabular}
\end{table*}

\begin{figure*}[t]
\centering
\begin{subfigure}[t]{0.75\textwidth}
\centering
\includegraphics[width=\textwidth]{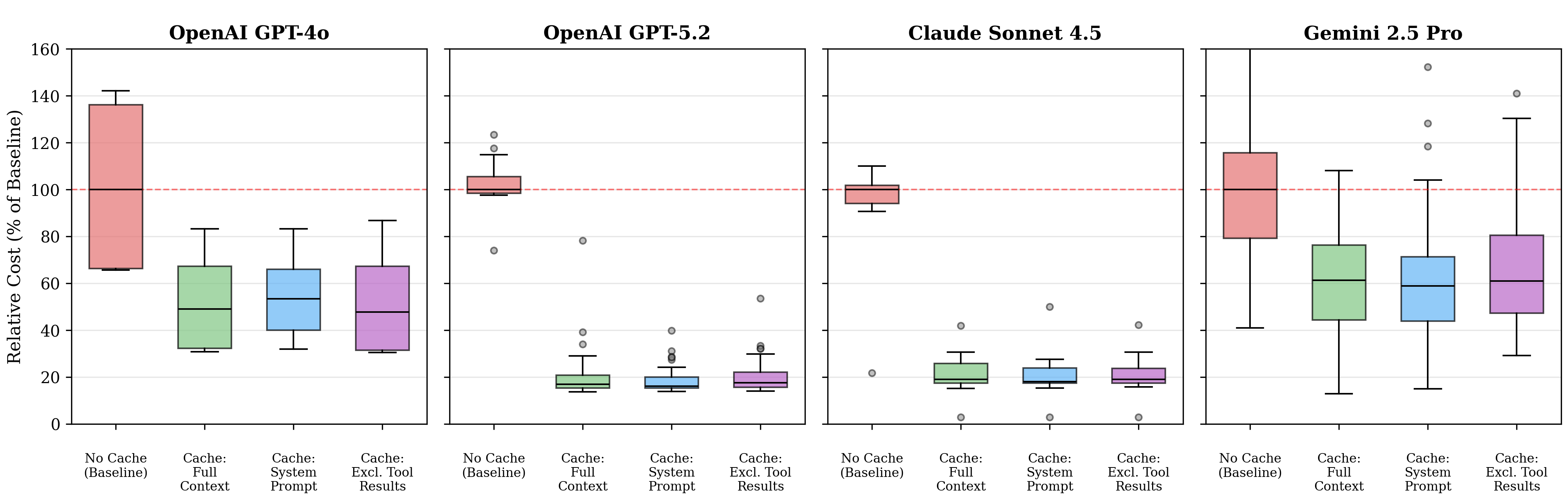}
\caption{Normalized cost distribution by cache mode}
\label{fig:cost_by_mode}
\end{subfigure}

\begin{subfigure}[t]{0.75\textwidth}
\centering
\includegraphics[width=\textwidth]{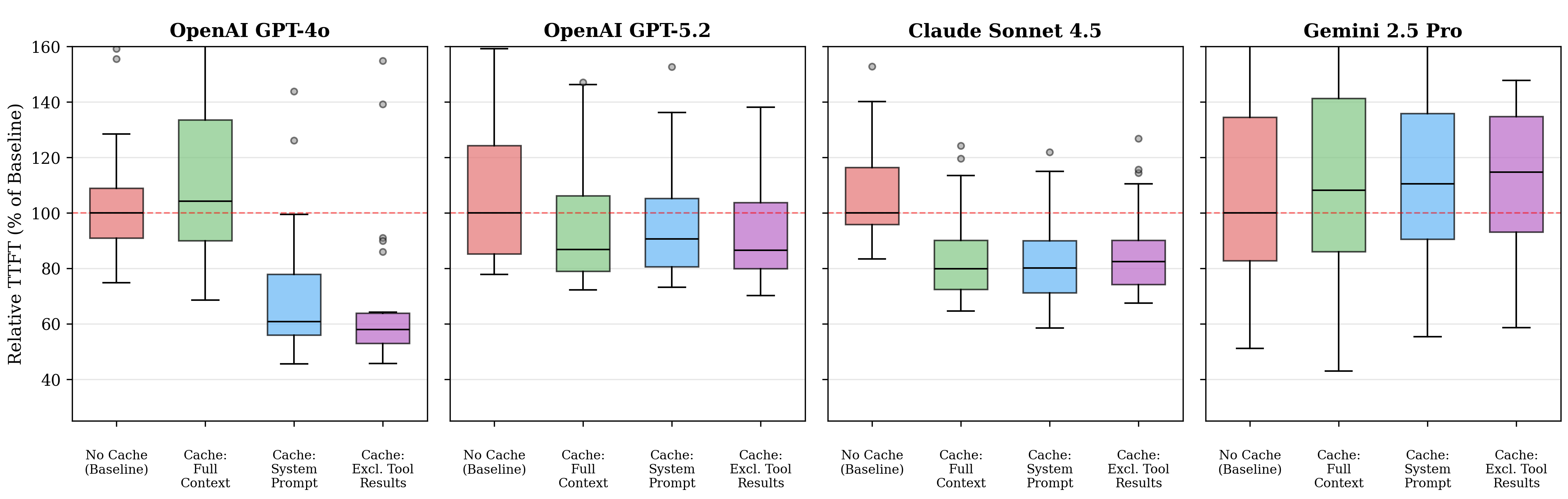}
\caption{Normalized TTFT distribution by cache mode}
\label{fig:ttft_by_mode}
\end{subfigure}
\caption{Normalized cost and TTFT distributions by model and cache strategy. Baseline = 100\%, lower is better. Cost savings are consistent across strategies - TTFT varies, with full-context caching sometimes underperforming selective strategies.}

\label{fig:cache_mode_comparison}
\end{figure*}

Prompt caching delivers substantial cost reductions across all providers and cache strategies. As shown in Table~\ref{tab:full_comparison} and Figure~\ref{fig:cost_ttft_comparison}, all cache modes achieve cost savings compared to the no-cache baseline for all four models. Cost reductions range from 79-81\% for GPT-5.2, 78-79\% for Claude Sonnet 4.5, 46-48\% for GPT-4o, and 28-41\% for Gemini 2.5 Pro depending on the cache mode selected. The consistency of cost savings across cache strategies suggests that the primary driver of cost reduction is caching the large system prompt, which remains stable across all requests within a session. Additional caching of conversation history and tool calls provides marginal incremental benefit for cost, as these components are smaller relative to the system prompt in our experimental setup.

Time to first token improvements show greater variation across providers compared to cost savings. GPT-4o shows 28-31\% improvement with system prompt only and exclude tool results strategies, while full context caching exhibits a slight regression of 8.8\%, suggesting that caching dynamic content can introduce overhead that negates latency benefits. Claude Sonnet 4.5 demonstrates consistent TTFT improvements across all cache strategies, ranging from 20.9\% to 22.9\%, including full context caching. This indicates that provider implementations differ in how they handle dynamic content caching. GPT-5.2 shows 13.0\% improvement with the exclude tool results strategy, while Gemini 2.5 Pro shows 6.1\% improvement with system prompt only caching. TTFT measurements exhibit natural variance due to factors including server load, network conditions, and provider infrastructure. This variance is reflected in the box plot distributions in Figure~\ref{fig:cache_mode_comparison}.

\subsection{Cache Strategy Comparison}
\label{subsec:strategy_comparison}

Figure~\ref{fig:cache_mode_comparison} presents normalized cost and TTFT distributions across all four cache strategies for each model. The results reveal important differences in how cache strategies perform across providers.

For cost optimization, all three caching strategies (full context, system prompt only, and exclude tool results) provide similar benefits within each model. The differences between strategies are typically within 2-4 percentage points, indicating that the system prompt, which is cached in all strategies, drives the majority of cost savings.

For latency optimization, the choice of cache strategy has a more pronounced impact. System prompt only caching and exclude tool results caching consistently outperform full context caching for TTFT improvement. For some models, full context caching shows no improvement or slight regression, while other strategies achieve 28-31\% improvement. The likely explanation is that full context caching triggers cache writes for dynamic tool calls and results, introducing overhead that offsets the benefits of cache reads.

\section{Discussion}
\label{sec:discussion}

\subsection{Strategic Cache Boundary Control}
\label{subsec:cache_boundaries}

Our results demonstrate that strategic control over cache boundaries is essential for maximizing prompt caching benefits in agentic workloads. The key insight is that providers abstract much of the underlying caching mechanism, automatically triggering cache creation when token thresholds are exceeded. Without explicit boundary control, this automatic behavior can cache dynamic, session-specific content that will not be reused, leading to cache write overhead without corresponding read benefits.

The most effective strategy is to ensure that only stable, reusable content is cached. In agentic applications, the system prompt is the most stable component, containing agent instructions, tool definitions, and persona guidelines that remain constant across sessions. Conversation history, tool calls, and tool results are dynamic and session-specific, making them poor candidates for cross-session caching. Practitioners should avoid including dynamic values in the system prompt itself. Common patterns that inadvertently break the cache include timestamps, datetime strings, session identifiers, or user-specific information embedded in the system prompt. If such dynamic information is necessary, it should be placed at the end of the system prompt to maximize the cacheable prefix. This ensures that the majority of the system prompt benefits from cache hits while only the dynamic suffix requires recomputation \cite{ji2025manus_context_engineering}.

%Similarly, dynamic function calling can break the cache when tool definitions change between requests. Modern agentic systems increasingly leverage dynamic tool discovery and registration through protocols such as the Model Context Protocol (MCP) \cite{modelcontextprotocol_intro}, where available tools may vary based on connected servers or runtime context \citep{lumer2025toolagentselectionsurvey}. When tool definitions are included in the prompt, any change to the available tool set invalidates the cached prefix. A practical strategy is to maintain a fixed set of general-purpose, reusable functions (such as code execution, file operations, and bash/shell commands), while implementing dynamic capabilities through code generation rather than traditional function calling \cite{ji2025manus_context_engineering,wang2024executablecodeactionselicit,anthropic_code_execution_with_mcp,anthropic2025_advanced_tool_use,cloudflare_code_mode,hn_code_mode_discussion}. Those prior methods of dynamic function calling, while achieving strong retrieval and execution accuracy, can prevent prompt caching usage \citep{lumer2024toolshedscaletoolequippedagents,lumer2025scalemcpdynamicautosynchronizingmodel,chen2024reinvoke,zheng2024toolrerank,chen2024reinvoke,wu2024sealtools}. 

Similarly, dynamic function calling can break the cache when tool definitions change between requests. Modern agentic systems increasingly leverage dynamic tool discovery through protocols such as the Model Context Protocol (MCP) \cite{modelcontextprotocol_intro}, where available tools may vary based on connected servers or runtime context \citep{lumer2025toolagentselectionsurvey}. When tool definitions are included in the prompt, any change to the available tool set invalidates the cached prefix. A practical strategy is to maintain a fixed set of general-purpose, reusable functions while implementing dynamic capabilities through code generation rather than traditional function calling \cite{ji2025manus_context_engineering,wang2024executablecodeactionselicit}. Prior methods of dynamic function calling can prevent prompt caching usage \citep{lumer2024toolshedscaletoolequippedagents,lumer2025scalemcpdynamicautosynchronizingmodel}.

\subsection{Tool Call Caching Considerations}
\label{subsec:tool_caching}

%For long-running agentic sessions with 30-50+ tool calls, practitioners may consider caching tool calls and results to reduce costs. However, this approach involves tradeoffs. Cache creation incurs a cost and latency overhead on the first request, which is only amortized if subsequent requests benefit from cache reads. For tool calls that produce highly variable results or that are unlikely to be repeated, caching provides no benefit and may introduce overhead.

%Common context management strategies in agentic systems can interact poorly with tool call caching. Techniques such as summarizing or pruning old tool calls to manage context length \citep{ji2025manus_context_engineering} inherently modify the conversation history, breaking any cached representations of that content. If an application employs such strategies, caching tool calls becomes counterproductive. The emerging pattern for agentic applications is to maintain a large, stable system prompt that benefits from caching, while treating tool calls and results as dynamic content that may be summarized, pruned, or otherwise managed throughout the session.

For long-running agentic sessions with 30-50+ tool calls, practitioners may consider caching tool calls and results to reduce costs. However, cache creation incurs overhead on the first request, which is only amortized if subsequent requests benefit from cache reads. For tool calls that produce variable results or are unlikely to be repeated, caching provides no benefit and may introduce overhead. 

Common context management strategies can interact poorly with tool call caching. Techniques such as summarizing or pruning old tool calls \citep{ji2025manus_context_engineering} break cached representations, making tool call caching counterproductive. The emerging pattern is to maintain a stable system prompt that benefits from caching while treating tool calls as dynamic content that may be managed throughout the session.

\subsection{Provider Implementation Variability}
\label{subsec:provider_variability}

Provider implementations of prompt caching differ in important ways that affect practical deployment. Minimum token thresholds for cache eligibility range from 1,024 to 4,096 tokens depending on the provider and model (see Appendix~\ref{app:pricing}). Time-to-live (TTL) durations vary from 5 minutes to 24 hours, affecting whether cached content remains available across user sessions. Some providers offer automatic caching that activates without developer intervention, while others require explicit API parameters to enable caching. Enterprise deployments may also leverage dedicated caching infrastructure \citep{liu2024cachegen,cheng2024large,10.1145/3689031.3696098,cheng2025lmcache} to further optimize performance. These implementation details are subject to change and practitioners should consult current provider documentation when designing caching strategies.

Our results reflect natural variance in API response times due to factors including server load, geographic distribution, and infrastructure differences across providers. When evaluating prompt caching benefits, practitioners should conduct experiments representative of their workloads and usage patterns rather than relying solely on published benchmarks. Practitioners should be aware of security considerations, as recent work has demonstrated that prompt caching can introduce timing side-channels that may leak information about cached content \citep{promptpeek2025, gu2025auditingpromptcachinglanguage}.

\section{Ablation Study}
\label{sec:ablation}

\begin{figure*}[!t]
\centering
\includegraphics[width=0.95\textwidth]{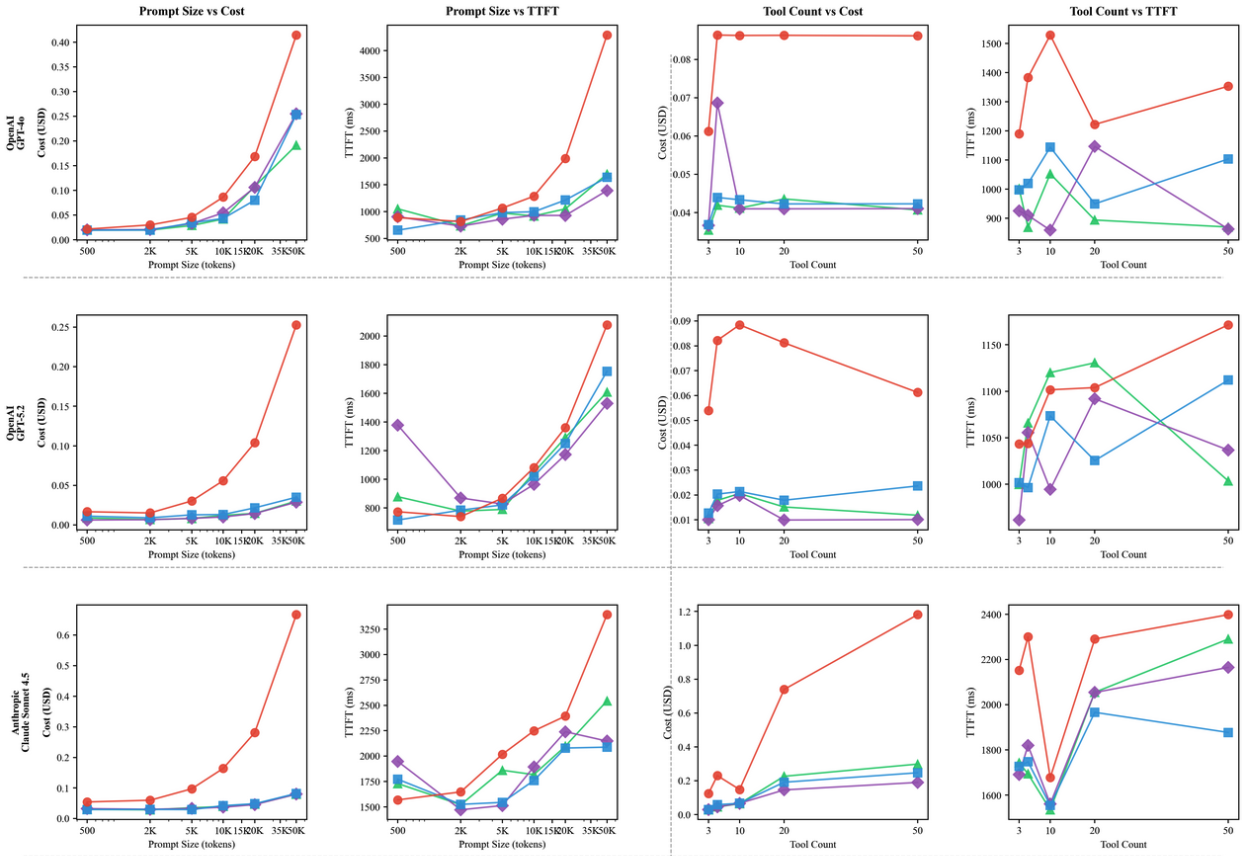}
\caption{Ablation results by prompt size (left) and tool count (right). Rows: GPT-4o, GPT-5.2, Claude Sonnet 4.5. Cost savings (columns 1, 3) are consistent across strategies vs.\ baseline (red). TTFT (columns 2, 4) varies more, below caching thresholds and at high counts.}\label{fig:ablation_grid}
\end{figure*}

To understand how prompt caching benefits scale with workload characteristics, we conduct an ablation study examining two key dimensions: prompt size and tool call count. These experiments isolate the factors that drive caching effectiveness and inform practitioners which workload characteristics benefit most from prompt caching. Figure~\ref{fig:ablation_grid} presents ablation results across three models and cache strategies.

\subsection{Ablation by Prompt Size}
\label{subsec:ablation_prompt_size}

We evaluate prompt caching across six prompt sizes: 500, 2,000, 5,000, 10,000, 20,000, and 50,000 tokens. For each size and model, we measure median cost and time to first token (TTFT) across all cache strategies. Notably, the 500-token condition falls below the minimum threshold required for prompt caching to activate (1,024 tokens for OpenAI and Anthropic, 4,096 tokens for Google; see Table~\ref{tab:caching_minimums}), serving as a control condition where caching cannot provide benefits.

As shown in the left columns of Figure~\ref{fig:ablation_grid}, cost savings scale linearly with prompt size across all models and are universally positive at every prompt size tested. At 50,000 tokens, GPT-5.2 achieves 89\% cost savings (from \$0.253 to \$0.029), Claude Sonnet 4.5 achieves 88\% savings (from \$0.667 to \$0.080), and GPT-4o achieves 54\% savings (from \$0.414 to \$0.192). TTFT improvements are also most pronounced at large context sizes, with GPT-4o showing 60\% improvement (from 4,290ms to 1,699ms) at 50,000 tokens.

At smaller prompt sizes (500-2,000 tokens), cost savings remain positive but modest, typically ranging from 10-45\%. However, TTFT results at small prompt sizes reveal an important nuance: at 500 tokens, which falls below the minimum caching threshold, GPT-4o, GPT-5.2, and Claude Sonnet 4.5 show TTFT regressions of 10-18\%. This is expected behavior, as the caching mechanism cannot activate below the threshold, and any observed latency differences reflect server variance rather than caching benefits. At prompt sizes above the threshold (2,000+ tokens), TTFT improvements become positive and scale with prompt size.

\subsection{Ablation by Tool Count}
\label{subsec:ablation_tool_count}

We evaluate prompt caching across five tool call counts: 3, 5, 10, 20, and 50 tool calls per session. For each configuration, we measure median cost and TTFT across cache strategies. The results (Figure~\ref{fig:ablation_grid}) demonstrate that cost savings remain consistent regardless of the number of tool calls in a session. GPT-5.2 maintains approximately 77-81\% cost savings across all tool counts, while GPT-4o achieves 42-53\% savings. This consistency indicates that the number of tool calls does not substantially affect the cost benefits of prompt caching. TTFT results for the tool count ablation show greater variance compared to the main study. While GPT-4o shows consistent TTFT improvements across all tool counts (16-36\%), other models exhibit mixed results. Claude Sonnet 4.5 shows diminishing TTFT returns at higher tool counts (from 19\% improvement at 3 tools to 5\% at 50 tools). This increased variance likely reflects server load fluctuations during the ablation experiments, as the main study (Section~\ref{sec:results}) showed more consistent TTFT improvements with lower variance.

\subsection{Discussion}
\label{subsec:ablation_discussion}

\textbf{Cost savings are universally positive.} Across all models, prompt sizes, and tool counts tested, prompt caching consistently reduces API costs. This finding provides practitioners with confidence that enabling prompt caching will reliably reduce costs regardless of workload characteristics.

\textbf{Prompt size drives caching benefits more than tool count.} Cost savings scale linearly with prompt size (from 10-45\% at 500 tokens to 54-89\% at 50,000 tokens), while remaining stable across tool counts (typically within 10 percentage points). This indicates the cacheable prefix length, primarily determined by the system prompt, is the dominant factor in caching effectiveness. Practitioners should focus on maximizing system prompt size within architectural constraints rather than optimizing around tool call patterns.

\textbf{TTFT improvements require meeting minimum thresholds.} The TTFT regressions at 500 tokens highlight the importance of meeting provider-specific thresholds (Table~\ref{tab:caching_minimums}). When prompts fall below these thresholds, caching cannot activate. The ablation study exhibits higher TTFT variance than the main study, reflecting the challenges of measuring latency in production API environments where server load and infrastructure changes introduce noise. Practitioners should conduct representative experiments for their workloads rather than relying on published benchmarks.

\section{Conclusion}
\label{sec:conclusion}

We present the first comprehensive evaluation of prompt caching for long-horizon agentic tasks across three major LLM providers. Our results demonstrate that prompt caching reduces API costs by 41 to 80\% and improves TTFT by 13 to 31\%. Strategic cache boundary control, such as caching only system prompts while excluding dynamic tool results, provides more consistent benefits than naive full context caching, which can paradoxically increase latency. These findings provide actionable guidance for deploying prompt caching in production agentic systems.

%\section*{Impact Statement}

%This paper evaluates prompt caching strategies to reduce the cost and latency of LLM-based agentic systems. The primary societal benefit is improving the efficiency and accessibility of AI systems by lowering operational costs, which may enable broader adoption and reduce energy consumption associated with redundant computation. We note that prompt caching can introduce timing side-channels that may leak information about cached content, a security consideration we discuss in Section~\ref{subsec:provider_variability}. Beyond this known concern, we do not foresee additional negative societal consequences that must be specifically highlighted.

\clearpage
\bibliography{references}
\bibliographystyle{icml2026}

\appendix
\newpage
\onecolumn
\section{Prompt caching pricing at time of evaluation}
\label{app:pricing}
\FloatBarrier

\begin{table}[ht]
\centering
\small
\setlength{\tabcolsep}{4pt}
\caption{
Token pricing as of early January 2026, used in our cost analysis (USD per 1M tokens unless otherwise noted).
``Cached input'' refers to cache-hit tokens. ``Cache write'' denotes cache creation costs when explicitly priced. Google additionally charges for context cache storage
(\$4.50 per million tokens per hour), which is accounted for separately in our analysis.
Pricing reflects public provider documentation at the time of evaluation. \citep{openai_prompt_caching_docs, anthropic_prompt_caching_docs, google_vertex_claude_prompt_caching_docs}
}
\label{tab:caching_pricing}
\begin{tabular}{
@{}p{2.6cm} p{3.6cm}
rrrr@{}
}
\toprule
\textbf{Provider} & \textbf{Model} &
\textbf{Input} &
\textbf{Output} &
\textbf{Cached} &
\textbf{Write} \\
\midrule
OpenAI
& GPT-4o
& 2.50 & 10.00 & 1.25 & --- \\

OpenAI
& GPT-5.2
& 1.75 & 14.00 & 0.175 & --- \\

\midrule
Anthropic
& Claude Sonnet 4.5
& 3.00 & 15.00 & 0.30 & 3.75 \\

\midrule
Google 
& Gemini 2.5 Pro ($\le$200K)
& 1.25 & 10.00 & 0.125 & --- \\

Google 
& Gemini 2.5 Pro ($>$200K)
& 2.50 & 15.00 & 0.250 & --- \\
\bottomrule
\end{tabular}

\vspace{0.4em}
\end{table}

\begin{table}[ht]
\centering
\small
\setlength{\tabcolsep}{4pt}
\caption{
Minimum prompt length (in tokens) required for prompt caching to apply, as of early January 2026.
Prompts shorter than these thresholds cannot benefit from caching, even when caching features are enabled.
Thresholds reflect public provider documentation at the time of evaluation. \citep{openai_prompt_caching_docs, anthropic_prompt_caching_docs, google_vertex_context_cache_overview}
}
\label{tab:caching_minimums}
\begin{tabular}{
@{}p{2.6cm} p{3.6cm}
r@{}
}
\toprule
\textbf{Provider} & \textbf{Model} &
\textbf{Min. Tokens} \\
\midrule
OpenAI
& GPT-4o
& 1,024 \\

OpenAI
& GPT-5.2
& 1,024 \\

\midrule
Anthropic
& Claude Sonnet 4.5
& 1,024 \\

\midrule
Google 
& Gemini 2.5 Pro
& 4,096 \\
\bottomrule
\end{tabular}

\vspace{0.4em}
\end{table}

\newpage
\section{Prompt Caching Mechanism}
\label{app:caching_mechanism}

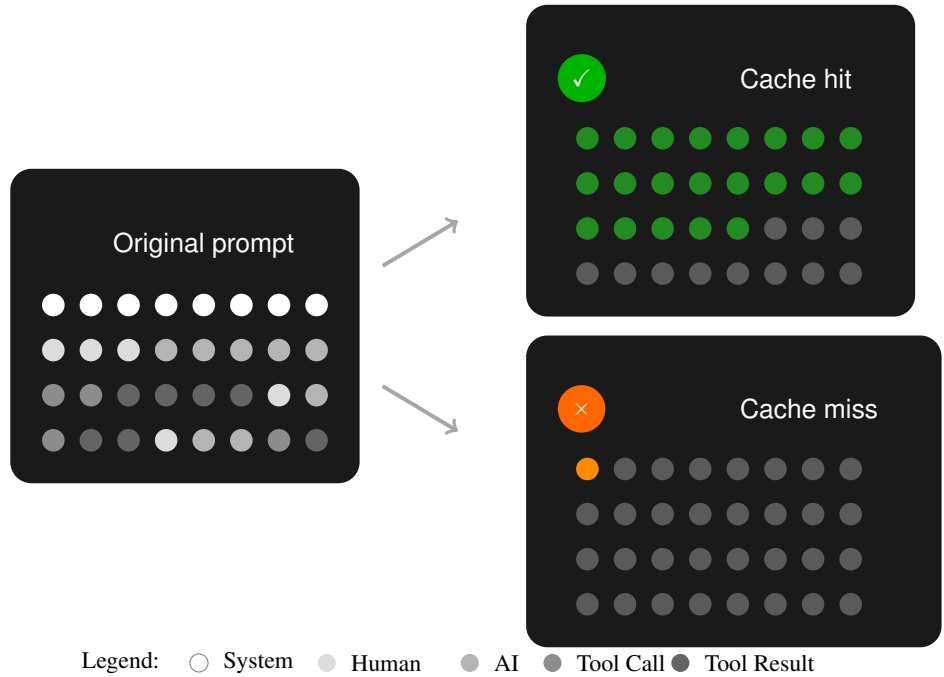
\begin{figure}[ht]
\centering
\begin{tikzpicture}[
    box/.style={rounded corners=8pt, fill=black!90, inner sep=12pt},
    boxtitle/.style={font=\sffamily, text=white},
    check/.style={circle, fill=green!70!black, text=white, font=\small, inner sep=3pt},
    cross/.style={circle, fill=orange!80!red, text=white, font=\small, inner sep=3pt}
]

% Message type shades (all circles, different brightness)
\definecolor{sysprompt}{RGB}{255,255,255}      % bright white - system prompt
\definecolor{humanmsg}{RGB}{220,220,220}       % light gray - human message  
\definecolor{aimsg}{RGB}{180,180,180}          % medium gray - AI message
\definecolor{toolcall}{RGB}{140,140,140}       % darker gray - tool call
\definecolor{toolresult}{RGB}{100,100,100}     % darkest gray - tool result

% Cache state colors
\definecolor{cached}{RGB}{34,139,34}           % green for cached
\definecolor{uncached}{RGB}{90,90,90}          % gray for uncached
\definecolor{mismatch}{RGB}{255,140,0}         % orange for mismatch

% Original prompt box
\node[box] (original) at (0,0) {
    \begin{tikzpicture}
        \node[boxtitle] at (2.0, 2.4) {Original prompt};
        
        % Row 1: System prompt (brightest)
        \foreach \i in {0,...,7} { \fill[sysprompt] (\i*0.5, 1.6) circle (0.15cm); }
        
        % Row 2: Human message + AI message
        \fill[humanmsg] (0, 1.0) circle (0.15cm);
        \fill[humanmsg] (0.5, 1.0) circle (0.15cm);
        \fill[humanmsg] (1.0, 1.0) circle (0.15cm);
        \fill[aimsg] (1.5, 1.0) circle (0.15cm);
        \fill[aimsg] (2.0, 1.0) circle (0.15cm);
        \fill[aimsg] (2.5, 1.0) circle (0.15cm);
        \fill[aimsg] (3.0, 1.0) circle (0.15cm);
        \fill[aimsg] (3.5, 1.0) circle (0.15cm);
        
        % Row 3: Tool call + Tool result
        \fill[toolcall] (0, 0.4) circle (0.15cm);
        \fill[toolcall] (0.5, 0.4) circle (0.15cm);
        \fill[toolresult] (1.0, 0.4) circle (0.15cm);
        \fill[toolresult] (1.5, 0.4) circle (0.15cm);
        \fill[toolresult] (2.0, 0.4) circle (0.15cm);
        \fill[toolresult] (2.5, 0.4) circle (0.15cm);
        \fill[humanmsg] (3.0, 0.4) circle (0.15cm);
        \fill[aimsg] (3.5, 0.4) circle (0.15cm);
        
        % Row 4: Mixed continuation
        \fill[toolcall] (0, -0.2) circle (0.15cm);
        \fill[toolresult] (0.5, -0.2) circle (0.15cm);
        \fill[toolresult] (1.0, -0.2) circle (0.15cm);
        \fill[humanmsg] (1.5, -0.2) circle (0.15cm);
        \fill[aimsg] (2.0, -0.2) circle (0.15cm);
        \fill[aimsg] (2.5, -0.2) circle (0.15cm);
        \fill[toolcall] (3.0, -0.2) circle (0.15cm);
        \fill[toolresult] (3.5, -0.2) circle (0.15cm);
    \end{tikzpicture}
};

% Arrows
\draw[->, line width=1.5pt, gray!70] (original.east) ++(0.3, 0.8) -- ++(1.0, 0.6);
\draw[->, line width=1.5pt, gray!70] (original.east) ++(0.3, -0.8) -- ++(1.0, -0.6);

% Cache hit box
\node[box, right=2.2cm of original, yshift=2.2cm] (cachehit) {
    \begin{tikzpicture}
        \node[check] at (-0.4, 2.4) {\checkmark};
        \node[boxtitle] at (1.6, 2.4) {Cache hit};
        
        % Row 1: System prompt - all cached (green)
        \foreach \i in {0,...,7} { \fill[cached] (\i*0.5, 1.6) circle (0.15cm); }
        
        % Row 2: Human + AI - all cached
        \foreach \i in {0,...,7} { \fill[cached] (\i*0.5, 1.0) circle (0.15cm); }
        
        % Row 3: Mostly cached, some new (gray)
        \fill[cached] (0, 0.4) circle (0.15cm);
        \fill[cached] (0.5, 0.4) circle (0.15cm);
        \fill[cached] (1.0, 0.4) circle (0.15cm);
        \fill[cached] (1.5, 0.4) circle (0.15cm);
        \fill[cached] (2.0, 0.4) circle (0.15cm);
        \fill[uncached] (2.5, 0.4) circle (0.15cm);
        \fill[uncached] (3.0, 0.4) circle (0.15cm);
        \fill[uncached] (3.5, 0.4) circle (0.15cm);
        
        % Row 4: New tokens (gray)
        \foreach \i in {0,...,7} { \fill[uncached] (\i*0.5, -0.2) circle (0.15cm); }
    \end{tikzpicture}
};

% Cache miss box
\node[box, right=2.2cm of original, yshift=-2.2cm] (cachemiss) {
    \begin{tikzpicture}
        \node[cross] at (-0.4, 2.4) {$\times$};
        \node[boxtitle] at (1.6, 2.4) {Cache miss};
        
        % Row 1: First token different (orange), rest gray
        \fill[mismatch] (0, 1.6) circle (0.15cm);
        \foreach \i in {1,...,7} { \fill[uncached] (\i*0.5, 1.6) circle (0.15cm); }
        
        % Row 2-4: All gray (recomputed)
        \foreach \i in {0,...,7} { \fill[uncached] (\i*0.5, 1.0) circle (0.15cm); }
        \foreach \i in {0,...,7} { \fill[uncached] (\i*0.5, 0.4) circle (0.15cm); }
        \foreach \i in {0,...,7} { \fill[uncached] (\i*0.5, -0.2) circle (0.15cm); }
    \end{tikzpicture}
};

% Legend - single row, spaced out
\node[below=2.0cm of original, xshift=3.5cm] (legend) {
    \begin{tikzpicture}
        \node[font=\footnotesize, anchor=east] at (-0.2, 0) {Legend:};
        
        \fill[sysprompt] (0.2, 0) circle (0.12cm);
        \draw[black!50] (0.2, 0) circle (0.12cm);
        \node[anchor=west, font=\footnotesize] at (0.4, 0) {System};
        
        \fill[humanmsg] (1.9, 0) circle (0.12cm);
        \node[anchor=west, font=\footnotesize] at (2.1, 0) {Human};
        
        \fill[aimsg] (3.8, 0) circle (0.12cm);
        \node[anchor=west, font=\footnotesize] at (4.0, 0) {AI};
        
        \fill[toolcall] (4.9, 0) circle (0.12cm);
        \node[anchor=west, font=\footnotesize] at (5.1, 0) {Tool Call};
        
        \fill[toolresult] (6.6, 0) circle (0.12cm);
        \node[anchor=west, font=\footnotesize] at (6.8, 0) {Tool Result};
    \end{tikzpicture}
};

\end{tikzpicture}
\caption{
Prompt caching requires exact prefix matches.
Different shades represent message types in agentic conversations: brightest (system prompt), light gray (human messages), medium gray (AI messages), darker gray (tool calls), and darkest (tool results).
Cache hit: The prompt prefix matches a previously seen request exactly, so cached KV tensors are reused (green) and only new tokens appended at the end require computation (gray).
Cache miss: Any difference in the prefix---even a single token at the beginning (orange)---prevents cache reuse, forcing full recomputation of all tokens (gray).
}
\label{fig:cache_mechanism}
\end{figure}

Figure~\ref{fig:cache_mechanism} illustrates the fundamental mechanism underlying prompt caching. When a request is processed, the system checks whether the prompt prefix matches previously cached content. A cache hit occurs when the entire prefix matches exactly, allowing the system to reuse previously computed KV tensors (shown in green). A cache miss occurs when any token differs from the cached content, even at the very beginning (shown with an orange indicator), forcing complete recomputation of all tokens (shown in gray).

%%%%%%%%%%%%%%%%%%%%%%%%%%%%%%%%%%%%%%%%%%%%%%%%%%%%%%%%%%%%%%%%%%%%%%%%%%%%%%%
% CACHE STRATEGY DIAGRAMS
%%%%%%%%%%%%%%%%%%%%%%%%%%%%%%%%%%%%%%%%%%%%%%%%%%%%%%%%%%%%%%%%%%%%%%%%%%%%%%%

\newpage
\section{Cache Strategy Implementations}
\label{app:cache_strategies}

The following figures illustrate the four cache strategies evaluated in this work. Since prompt caching operates on exact prefix matches, we use UUIDs (indicated by red bars) to control cache boundaries. Static content placed before the UUID forms the cacheable prefix; content after the UUID varies between requests and prevents prefix matches beyond that point.

% Shared colors for all strategy figures
\definecolor{syspromptc}{RGB}{255,255,255}
\definecolor{humanmsgc}{RGB}{220,220,220}
\definecolor{aimsgc}{RGB}{180,180,180}
\definecolor{toolcallc}{RGB}{140,140,140}
\definecolor{toolresultc}{RGB}{100,100,100}
\definecolor{uuidcolor}{RGB}{220,50,50}
\definecolor{cacheablegreen}{RGB}{34,139,34}
\definecolor{notcachedgray}{RGB}{90,90,90}

%--- NO CACHE (BASELINE) ---
\subsection{No Cache (Baseline)}
\label{app:no_cache}

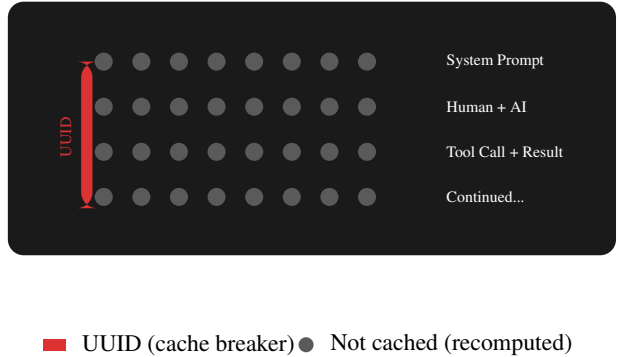
\begin{figure}[ht]
\centering
\begin{tikzpicture}[
    box/.style={rounded corners=6pt, fill=black!90, inner sep=10pt}
]

\node[box] (prompt) at (0,0) {
    \begin{tikzpicture}
        % UUID at the very beginning
        \fill[uuidcolor] (-0.3, 1.6) rectangle (-0.15, -0.35);
        \node[font=\tiny, text=uuidcolor, rotate=90] at (-0.5, 0.6) {UUID};
        
        % Row 1: System prompt (all gray - not cached)
        \foreach \i in {0,...,7} { \fill[notcachedgray] (\i*0.5, 1.6) circle (0.12cm); }
        \node[font=\tiny, text=white, anchor=west] at (4.2, 1.6) {System Prompt};
        
        % Row 2: Human + AI (gray)
        \foreach \i in {0,...,2} { \fill[notcachedgray] (\i*0.5, 1.0) circle (0.12cm); }
        \foreach \i in {3,...,7} { \fill[notcachedgray] (\i*0.5, 1.0) circle (0.12cm); }
        \node[font=\tiny, text=white, anchor=west] at (4.2, 1.0) {Human + AI};
        
        % Row 3: Tool call + result (gray)
        \foreach \i in {0,...,1} { \fill[notcachedgray] (\i*0.5, 0.4) circle (0.12cm); }
        \foreach \i in {2,...,7} { \fill[notcachedgray] (\i*0.5, 0.4) circle (0.12cm); }
        \node[font=\tiny, text=white, anchor=west] at (4.2, 0.4) {Tool Call + Result};
        
        % Row 4: More conversation (gray)
        \foreach \i in {0,...,7} { \fill[notcachedgray] (\i*0.5, -0.2) circle (0.12cm); }
        \node[font=\tiny, text=white, anchor=west] at (4.2, -0.2) {Continued...};
    \end{tikzpicture}
};

% Legend
\node[below=0.8cm of prompt] {
    \begin{tikzpicture}
        \fill[uuidcolor] (0, 0) rectangle (0.3, 0.15);
        \node[anchor=west, font=\footnotesize] at (0.4, 0.075) {UUID (cache breaker)};
        \fill[notcachedgray] (3.5, 0.075) circle (0.1cm);
        \node[anchor=west, font=\footnotesize] at (3.7, 0.075) {Not cached (recomputed)};
    \end{tikzpicture}
};

\end{tikzpicture}
\caption{No Cache (Baseline): A unique UUID prepended to the start of the system prompt ensures no prefix match is possible with any prior request, forcing full recomputation of all tokens every time.}
\label{fig:no_cache_strategy}
\end{figure}

%--- FULL CONTEXT CACHING ---
\subsection{Full Context Caching}
\label{app:full_context}

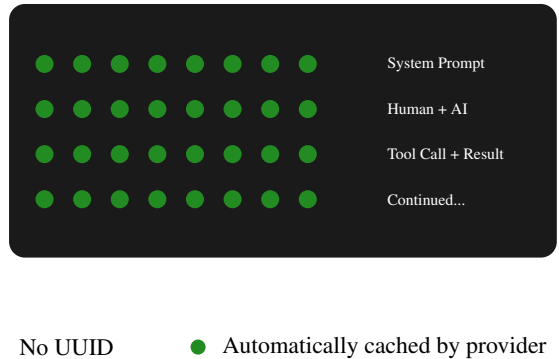
\begin{figure}[ht]
\centering
\begin{tikzpicture}[
    box/.style={rounded corners=6pt, fill=black!90, inner sep=10pt}
]

\node[box] (prompt) at (0,0) {
    \begin{tikzpicture}
        % No UUID - everything can be cached
        
        % Row 1: System prompt (green - cached)
        \foreach \i in {0,...,7} { \fill[cacheablegreen] (\i*0.5, 1.6) circle (0.12cm); }
        \node[font=\tiny, text=white, anchor=west] at (4.2, 1.6) {System Prompt};
        
        % Row 2: Human + AI (green - cached)
        \foreach \i in {0,...,7} { \fill[cacheablegreen] (\i*0.5, 1.0) circle (0.12cm); }
        \node[font=\tiny, text=white, anchor=west] at (4.2, 1.0) {Human + AI};
        
        % Row 3: Tool call + result (green - cached)
        \foreach \i in {0,...,7} { \fill[cacheablegreen] (\i*0.5, 0.4) circle (0.12cm); }
        \node[font=\tiny, text=white, anchor=west] at (4.2, 0.4) {Tool Call + Result};
        
        % Row 4: More conversation (green - cached)
        \foreach \i in {0,...,7} { \fill[cacheablegreen] (\i*0.5, -0.2) circle (0.12cm); }
        \node[font=\tiny, text=white, anchor=west] at (4.2, -0.2) {Continued...};
    \end{tikzpicture}
};

% Legend
\node[below=0.8cm of prompt] {
    \begin{tikzpicture}
        \node[anchor=west, font=\footnotesize] at (0, 0.075) {No UUID};
        \fill[cacheablegreen] (2.5, 0.075) circle (0.1cm);
        \node[anchor=west, font=\footnotesize] at (2.7, 0.075) {Automatically cached by provider};
    \end{tikzpicture}
};

\end{tikzpicture}
\caption{Full Context Caching: No UUIDs are added, allowing the provider to automatically cache the entire prompt prefix. However, this may cache dynamic content (e.g., tool results) that varies between sessions, potentially triggering cache writes without corresponding cache hits.}
\label{fig:full_context_strategy}
\end{figure}

%--- SYSTEM PROMPT ONLY ---
\newpage
\subsection{System Prompt Only Caching}
\label{app:system_prompt_only}

\begin{figure}[ht]
\centering
\begin{tikzpicture}[
    box/.style={rounded corners=6pt, fill=black!90, inner sep=10pt}
]

\node[box] (prompt) at (0,0) {
    \begin{tikzpicture}
        % Row 1: System prompt (green - cached)
        \foreach \i in {0,...,7} { \fill[cacheablegreen] (\i*0.5, 1.6) circle (0.12cm); }
        \node[font=\tiny, text=white, anchor=west] at (4.2, 1.6) {System Prompt};
        
        % UUID after system prompt
        \fill[uuidcolor] (-0.1, 1.35) rectangle (4.1, 1.25);
        \node[font=\tiny, text=uuidcolor] at (5.0, 1.3) {UUID};
        
        % Row 2: Human + AI (gray - not cached)
        \foreach \i in {0,...,7} { \fill[notcachedgray] (\i*0.5, 1.0) circle (0.12cm); }
        \node[font=\tiny, text=white, anchor=west] at (4.2, 1.0) {Human + AI};
        
        % Row 3: Tool call + result (gray - not cached)
        \foreach \i in {0,...,7} { \fill[notcachedgray] (\i*0.5, 0.4) circle (0.12cm); }
        \node[font=\tiny, text=white, anchor=west] at (4.2, 0.4) {Tool Call + Result};
        
        % Row 4: More conversation (gray - not cached)
        \foreach \i in {0,...,7} { \fill[notcachedgray] (\i*0.5, -0.2) circle (0.12cm); }
        \node[font=\tiny, text=white, anchor=west] at (4.2, -0.2) {Continued...};
    \end{tikzpicture}
};

% Legend
\node[below=0.8cm of prompt] {
    \begin{tikzpicture}
        \fill[uuidcolor] (0, 0) rectangle (0.3, 0.15);
        \node[anchor=west, font=\footnotesize] at (0.4, 0.075) {UUID};
        \fill[cacheablegreen] (1.8, 0.075) circle (0.1cm);
        \node[anchor=west, font=\footnotesize] at (2.0, 0.075) {Cached};
        \fill[notcachedgray] (3.3, 0.075) circle (0.1cm);
        \node[anchor=west, font=\footnotesize] at (3.5, 0.075) {Not cached};
    \end{tikzpicture}
};

\end{tikzpicture}
\caption{System Prompt Only Caching: A UUID appended after the system prompt breaks the cacheable prefix at this boundary. The static system prompt (placed at the beginning) benefits from prefix caching, while dynamic conversation content (placed after) is recomputed each request.}
\label{fig:system_prompt_only_strategy}
\end{figure}
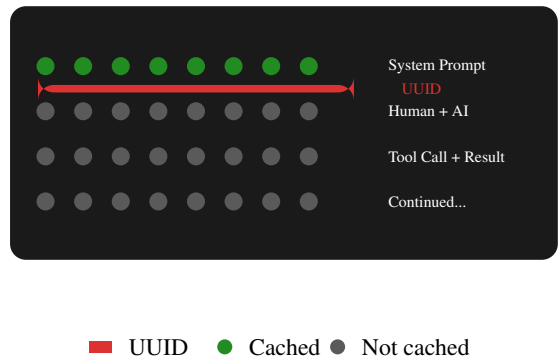

%--- EXCLUDE TOOL RESULTS ---
\subsection{Exclude Tool Results Caching}
\label{app:exclude_tool_results}

\begin{figure}[ht]
\centering
\begin{tikzpicture}[
    box/.style={rounded corners=6pt, fill=black!90, inner sep=10pt}
]

\node[box] (prompt) at (0,0) {
    \begin{tikzpicture}
        % Row 1: System prompt (green - cached)
        \foreach \i in {0,...,7} { \fill[cacheablegreen] (\i*0.5, 1.6) circle (0.12cm); }
        \node[font=\tiny, text=white, anchor=west] at (4.2, 1.6) {System Prompt};
        
        % UUID after system prompt
        \fill[uuidcolor] (-0.1, 1.35) rectangle (4.1, 1.25);
        \node[font=\tiny, text=uuidcolor] at (5.0, 1.3) {UUID};
        
        % Row 2: Human + AI (gray - not cached due to UUID above)
        \foreach \i in {0,...,7} { \fill[notcachedgray] (\i*0.5, 1.0) circle (0.12cm); }
        \node[font=\tiny, text=white, anchor=west] at (4.2, 1.0) {Human + AI};
        
        % Row 3: Tool call (gray)
        \foreach \i in {0,...,1} { \fill[notcachedgray] (\i*0.5, 0.55) circle (0.12cm); }
        \node[font=\tiny, text=white, anchor=west] at (1.2, 0.55) {Tool Call};
        
        % Row 4: Tool result (gray)
        \foreach \i in {0,...,5} { \fill[notcachedgray] (\i*0.5, 0.2) circle (0.12cm); }
        \node[font=\tiny, text=white, anchor=west] at (3.2, 0.2) {Tool Result};
        
        % UUID after tool result
        \fill[uuidcolor] (-0.1, -0.05) rectangle (4.1, -0.15);
        \node[font=\tiny, text=uuidcolor] at (5.0, -0.1) {UUID};
        
        % Row 5: Next turn (gray)
        \foreach \i in {0,...,7} { \fill[notcachedgray] (\i*0.5, -0.4) circle (0.12cm); }
        \node[font=\tiny, text=white, anchor=west] at (4.2, -0.4) {Next Human + AI};
        
        % Row 6: Another tool call
        \foreach \i in {0,...,1} { \fill[notcachedgray] (\i*0.5, -0.75) circle (0.12cm); }
        \node[font=\tiny, text=white, anchor=west] at (1.2, -0.75) {Tool Call};
        
        % Row 7: Another tool result
        \foreach \i in {0,...,5} { \fill[notcachedgray] (\i*0.5, -1.1) circle (0.12cm); }
        \node[font=\tiny, text=white, anchor=west] at (3.2, -1.1) {Tool Result};
        
        % UUID after second tool result
        \fill[uuidcolor] (-0.1, -1.35) rectangle (4.1, -1.45);
        \node[font=\tiny, text=uuidcolor] at (5.0, -1.4) {UUID};
    \end{tikzpicture}
};

% Legend
\node[below=0.8cm of prompt] {
    \begin{tikzpicture}
        \fill[uuidcolor] (0, 0) rectangle (0.3, 0.15);
        \node[anchor=west, font=\footnotesize] at (0.4, 0.075) {UUID (after system + each tool result)};
        \fill[cacheablegreen] (5.8, 0.075) circle (0.1cm);
        \node[anchor=west, font=\footnotesize] at (6.0, 0.075) {Cached};
        \fill[notcachedgray] (7.3, 0.075) circle (0.1cm);
        \node[anchor=west, font=\footnotesize] at (7.5, 0.075) {Not cached};
    \end{tikzpicture}
};

\end{tikzpicture}
\caption{Exclude Tool Results Caching: UUIDs are appended after the system prompt and after each tool result to break the cacheable prefix at these boundaries. This prevents session-specific tool results from being cached, avoiding cache writes for content unlikely to produce future cache hits. Furthermore, this mirrors cache-breaking context engineering strategies that prune or summarize past tool calls.}
\label{fig:exclude_tool_results_strategy}
\end{figure}
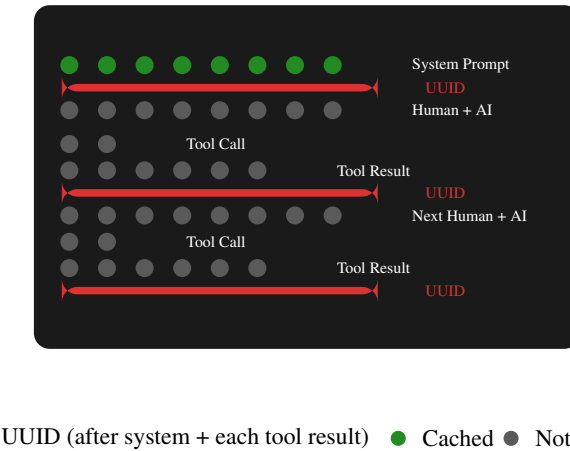

\end{document}